\pdfoutput=1
\documentclass[11pt]{article}

\usepackage{acl}

\usepackage{times}
\usepackage{latexsym}

\usepackage[T1]{fontenc}
\usepackage[utf8]{inputenc}

\usepackage{microtype}
\usepackage{graphicx}
\usepackage{booktabs}

\usepackage{algorithm,algpseudocode}
\title{Easy and Efficient Transformer: Scalable Inference Solution For Large NLP Model}

\author{Gongzheng Li$^{1}$\thanks{\quad Equal contribution}, Yadong Xi$^1$\footnotemark[1], Jingzhen Ding$^1$, Duan Wang$^1$, Ziyang Luo$^2$,\\{\bf Rongsheng Zhang$^1$,Bai Liu$^1$, Changjie Fan$^1$, Xiaoxi Mao$^1$\thanks{\quad Corresponding Author}, Zeng Zhao$^1$\footnotemark[2]}\\
  $^1$ {\normalsize Fuxi AI Lab, NetEase Inc., Hangzhou, China} \\
  $^2$  {\normalsize Department of Computer Science, Hong Kong Baptist University, Hong Kong SAR, China} \\
  \texttt{\normalsize \{ligongzheng,xiyadong,maoxiaoxi,hzzhaozeng\}@corp.netease.com}}

\begin{document}
\maketitle
\begin{abstract}

  Recently, large-scale transformer-based models have been proven to be effective over various tasks across many domains. Nevertheless, applying them in industrial production requires tedious and heavy works to reduce inference costs. To fill such a gap, we introduce a scalable inference solution: \textbf{Easy and Efficient Transformer (EET)}, including a series of transformer inference optimization at the algorithm and implementation levels. First, we design highly optimized kernels for long inputs and large hidden sizes. Second, we propose a flexible CUDA memory manager to reduce the memory footprint when deploying a large model. Compared with the state-of-the-art transformer inference library (Faster Transformer v4.0), EET can achieve an average of 1.40-4.20x speedup on the transformer decoder layer with an A100 GPU.

\end{abstract}

\section{Introduction}

% Background introduction
In recent years, transformer-based models have achieved impressive performance across variant domains, such as natural language processing~\cite{vaswani2017attention,devlin2019bert,2020t5,brown2020language}, computer vision~\cite{jiang2021transgan,dosovitskiy2020image} and speech processing~\cite{baevski2020wav2vec,baevski2021unsupervised}. The scaling law proposed by~\citet{kaplan2020scaling} indicates that the validation PPL of a neural language model scales as a power-law with model sizes, dataset sizes, and the amount of training computation. Such law is corroborated empirically by many following works~\cite{brown2020language,zhai2021scaling}.

% importance of our works
However, the mega-sized models are notoriously expensive for deployment in the industry. For example, GPT-2 medium model (700M parameters~\cite{radford2019language}) spends up to 10s to generate 512 tokens given a prompt with the length of 512 on an RTX 2080ti GPU, which is not allowed in the industrial application. Multiple approaches have been proposed to solve such problems, including knowledge distillation~\cite{hinton2015distilling,jiao2020tinybert}, model pruning~\cite{voita2019analyzing}, and quantization~\cite{shen2019qbert}.  Apart from these works, much attention has also been paid to optimizing CUDA implementation of a transformer layer for better hardware utilization. Previous works (e.g.: TensorRT~\cite{Tensorrt-2021}, LightSeq~\cite{wang2021lightseq} and Faster Transformer (FT) ~\cite{Faster-Transformer-V3.1-2021}) have implemented many optimization techniques, including kernels fusion, gemm optimization, quantization, etc. However, these works still have several limitations. TensorRT only contains the multi-head attention(MHA) operation, lacking a complete transformer model. LightSeq cannot support the model hidden size and input sequence length over 1024. FT contains some performance flaws which need to be improved.

In this paper, we propose a novel transformer inference acceleration library, \textbf{Easy and Efficient Transformer (EET) }. First, we implement custom CUDA kernels to avoid explicit matrix addition of attention and padding masks with attention weights. As a result, the attention mask matrix is no longer required, while FT spends overhead to initialize an attention mask on the CPU and push it to CUDA. In addition, compared with FT, padding masks are no longer needed in computation, leading to additional performance improvement. Second, we propose a new method, \textit{thread block folding}, to extend all kernels to support a larger model size up to 12288 and a longer sequence up to 4096. For FT, it directly assigns the thread number in a CUDA block, which may hurt the parallel efficiency. Finally, we design a dynamic CUDA memory management mechanism to reduce the CUDA memory occupation for the same model size, while FT needs to manually allocate memory usage.

We have conducted comprehensive experiments to compare EET with Fairseq,\footnote{https://github.com/pytorch/fairseq} LightSeq and FT. In our experiments, EET achieves about 4.48-20.27x and 4.30-27.43x speedup over Fairseq on 2080ti and A100 respectively. When we set the model size to 768 and 1024 on 2080Ti, EET makes 0.82-2.46x speedup over LightSeq. Compared to FT(v3.1), EET achieves about 1.21-6.30x and 1.62-8.16x speedup on 2080ti and A100 respectively. Compared to FT(v4.0), EET achieves about 1.40-4.20x speedup on A100. The remarkable experimental results corroborate the effectiveness of our EET.

\section{Custom Kernels}
FT~\cite{Faster-Transformer-V3.1-2021} has implemented highly optimized CUDA kernels for transformer inference. To make further optimization, we design our custom kernels with the considerations below:

$\bullet$ Because padding tokens do not affect the final results, preventing padding tokens from participating in MHA instead of simply applying padding masks can significantly reduce the computational overhead.

$\bullet$ Although an attention mask is essential for MHA in text generation, constructing a mask that varies with the input length is time-consuming.

$\bullet$ The hidden sizes and input lengths of the large-scale pre-trained models can easily exceed 1024. It is necessary to extend these kernels to support large hidden sizes and input lengths elegantly and efficiently.  

To remove previously mentioned masks in computation, we redesign the kernels and call the mechanism \textit{mask fusion}. To extend all the kernels to support the model size or sequence length greater than 1024, we improve the CUDA thread structure and call the method as \textit{thread block folding}. Next, we describe these two methods in detail.

\subsection{Mask Fusion}
 
The attention mask indicates the attention boundary for each token to prevent the
attention from looking forward. The padding mask indicates where the padding tokens are. Thus they both characterize the position information of the tokens in a sequence. Meanwhile, each CUDA thread also has a unique positional index. So we can map each token in the MHA to a thread or block in the CUDA kernels. The function of the attention mask is achieved by comparing whether the CUDA position of the query token being processed is larger than the CUDA position of the key token. The function of the padding mask is achieved by starting the valid calculations from the padding offset when sequentially processing each token. Therefore, we transform the mask computation to logical operation with CUDA thread index comparison. Thus there is no need to store any explicit functional parameters of the masks and the computation overhead of masking operation is saved. The algorithm pseudo-code is shown in Algorithm~\ref{alg:gpt-2 attention}. 

\begin{algorithm}
  \small
  \caption{MHA with \textit{mask fusion}}
  \label{alg:gpt-2 attention}
\begin{algorithmic}
  \State {\bfseries Input:} $qk, paddingLen, seqLen, batch, headNum$  
  \State {\bfseries Output:} the attention weights back to $qk$
  \State CUDA Initialize $grid \gets (batch * headNum)$
  \State CUDA Initialize $ block \gets (seqLen)$
  \State $batchId \gets blockIdx.x / headNum$
  \State $padLen \gets paddingLen[batchId]$
  \State $qkOffset \gets blockIdx.x * seqLen * seqLen$
  \State $qkOffset \gets qkOffset + paddLen * seqLen$
  \State $s \gets padLen$ \algorithmiccomment{\textcolor{red}{start at first non-pad}}
  \State $e \gets seqLen$ \algorithmiccomment{\textcolor{red}{end at last token}}
  \State $reduceMax \gets -inf$
  \State $reduceSum \gets 0$
  \For{$i=s$ {\bfseries to} $e$}
  \State $position \gets qkOffset + threadIdx.x$
  \State $data \gets qk[position]$
  \State $u \gets padLen$ \algorithmiccomment{\textcolor{red}{upper boundary}}
  \State $l \gets i$  \algorithmiccomment{\textcolor{red}{lower boundary}}
  \If{$l <threadIdx.x < u$}
  \State $reduceMax \gets blockReduceMax(data)$
  \State $reduceSum \gets blockReduceSum(data)$
  \State $data \gets softmax(reduceMax, reduceSum)$
  \EndIf
  \State $qk[position] \gets data$
  \EndFor
\end{algorithmic}
\end{algorithm}

\subsection{Thread Block Folding}
Large-scale models often have model sizes and input lengths larger than 1024. For example, the standard GPT-3 has a model size of 12288 and an input length of 2048. However, the CUDA block only supports a maximum thread number of 1024, most inference frameworks, such as FT(v3.1) and LightSeq, have implemented kernels that restrict the model size and input length up to 1024, leading to limited availability. 

\begin{figure*}[htbp]
  \centering
  \includegraphics[width=1.0\textwidth]{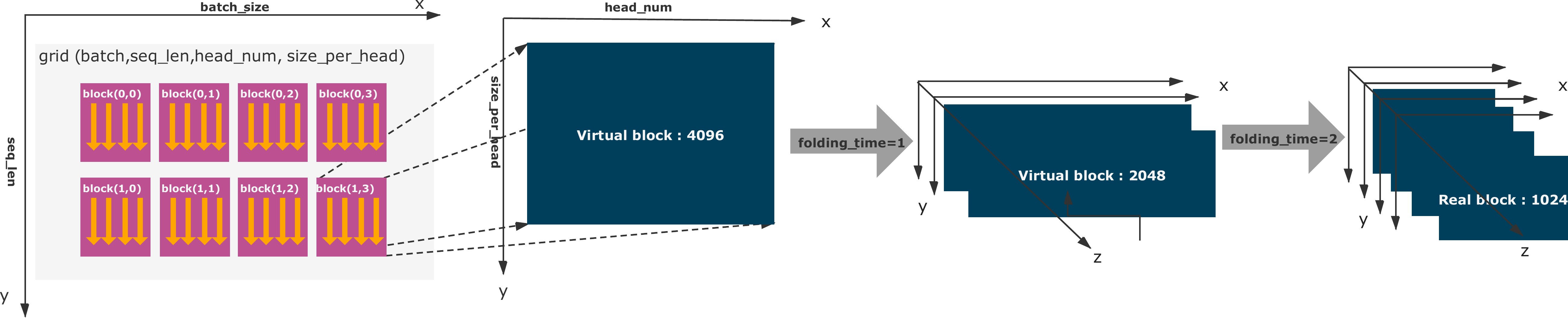}
  \caption{The schematic diagram of \textit{thread block folding}.}
  \label{figure-thread-fold}
\end{figure*}

To deal with large model sizes and sequence lengths, we propose to use several blocks to simulate a large block, shown as Figure~\ref{figure-thread-fold}. Imagine a virtual block large enough to hold all the tasks, then we can fold it once to create two blocks, with each block having half the size of the original block. We can repeat the process until the sub-blocks size satisfies the CUDA constraint. Then, the large model sizes or input lengths can be handled correctly, and a new CUDA thread dimension is created to manage the folding procedure. We call this method \textit{thread block folding},  which allows us to extend any kernel to any model size and any sequence length with minimum changes and non-degraded performance. For instance, assuming the model size is 1280, we fold it once and create two half-size blocks, then the data can be assigned into two separate blocks with 640 threads in each.

We introduce a folding coefficient to characterize the number of folding. Given the model size $h$, the folding coefficient $t$ and the number of threads $n$ in one block is defined as: 
\[ t = 2 ^ {\lceil \frac{h}{1024} \rceil - 1}; \quad n = \frac{h}{2^{t}} \]
As for simplicity, \textit{thread block folding} only adds a new dimension for the block, which slightly impacts the basic CUDA thread grid structure. As for efficiency, the minimum thread number is 512 when the model size or input length is larger than 1024 and makes full use of thread parallelism. The sequence expansion process is similar to the model expansion process. Finally, we support the model size no larger than 16384 and sequence length no longer than 4096.

\section{Dynamic Memory Manager}
The inference is much more sensitive to latency compared to training. As a result, model parallelism~\cite{shoeybi2020megatronlm} and pipeline parallelism~\cite{huang2019gpipe} are undesirable for inference. Their communication overhead introduced by tensor slicing or layer split is significant even with the support of NVLink and GPUDirect. To reduce the latency and hardware requirements for online service, minimizing the memory footprint is of exceptional value when loading very large models. Thus we propose a dynamic memory management strategy for this issue.

Except for the model weights, the memory footprint includes the caches and the buffers. It is hard to reduce the memory footprint of weights because they are inherent to the model. Similarly, The $K$/$V$ caches for MHA are also hard to compress because they are pre-allocated to avoid runtime memory requests, the size of which depends on the model size, maximum batch size, and maximum sequence length. Whereas the activation cache and the buffers used to store the operator's results are compressible. Hence our dynamic memory management strategy mainly focuses on the activation caches and the operation result buffers.

\subsection{Cache Reuse}
The caches include $K$/$V$ caches and activation caches. In incremental decoding, the keys and the values for every step are stored for the next step's attention computation. The maximum size of $K$/$V$ caches is predictable because we can determine the maximum batch size and decoding steps at the start of the running instance. We allocate the maximum required memory in advance to reduce the forward latency, avoiding malloc overhead and memory corruption. 

Different from $K$/$V$ caches, the activation results are useless after we have calculated and passed them to the next layer. The memory for these activations can be reused across different layers and different operators. We could reuse the activation caches in the following cases. 

$\bullet$ The embedding operator shares the cache with the feed-forward operator and the final output. Yet the attention operator holds another cache because of the residual connection. 

$\bullet$ The cache for input sequences can be reused by the decoded tokens. The maximum size is determined by the maximum input length. 

$\bullet$ The cache can be reused across different layers.

We use the following notations: $b$, the maximum batch size; $s$, the maximum sequence length; $p$, the maximum prompt length; $h$, the hidden units; $l$, the layer number.
 The total activation cache size is:
\[ 2 * b * h * p \]
 The total K/V cache size is :
 \[2 * b * h * s * l \]

 \subsection{Buffer Reuse}
The continuous CUDA kernels are not always fused, especially when it comes to Cublas GEMM calls. So we need some buffers to store the returns for those non-fused kernels. Managing the buffers manually like FT is complicated and inefficient. We develop a dynamic buffer manager to avoid the tedium of manual design and achieve a highly efficient memory allocation.

\begin{figure}[htbp]
  \centering
  \includegraphics[width=0.45\textwidth]{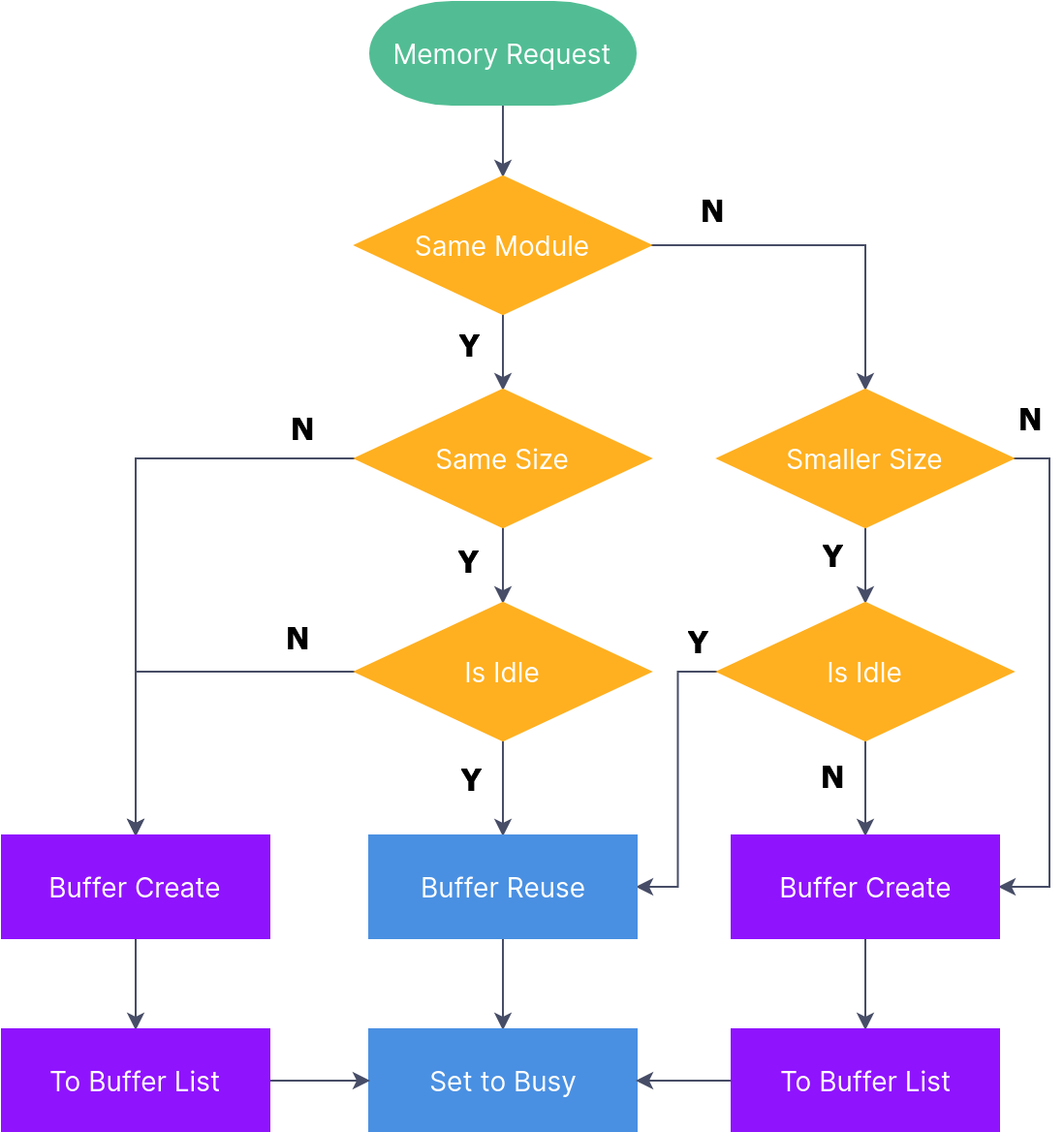}
  \caption{The schematic diagram of buffer decision strategy.}
  \label{figure buffer_tree}
\end{figure}

We maintain a list of buffers and use different strategies within and across modules to improve memory utilization. When within modules, we reuse the buffer only when the request size is identical to an idle buffer in the list, preventing memory fragmentation. When across modules, we reuse the buffer when the request size is smaller than any idle buffer in the list, avoiding duplicated malloc. The decision process is demonstrated in Figure \ref{figure buffer_tree}. In our design, the developer only needs to request a buffer of a specified size and mark it as idle when it is useless, without concerning how to reuse memory exactly. The total buffer size is:
\[ b * p * (6 * h + n * p) \]
where $b$ is the batch size, $p$ is the input length, $h$ is the hidden size, and $n$ is the head number.

\section{Experiments}
During inference, many factors can affect the actual performance, including model size, prompt length, sequence length, padding ratio in a batch, and hardware feature. Completely traversing all combinations requires a huge amount of works. Because the dataset has no effect on the experiment results, we adopt the fake inputs for convenience. To compare fairly and reduce our works, we define some typical experiment settings. If there is no special instruction, the experiment is conducted based on Configuration A in Table~\ref{tabel configuration}. Fairseq is an intuitional baseline because it is implemented using pure PyTorch code.

\begin{table}[htbp]
  \caption{Configuration A and B}
  \label{tabel configuration}
  \begin{center}
  \begin{small}
  \begin{sc}
  \begin{tabular}{lcccr}
  \toprule
   & Config A & Config B\\
  \midrule
  Batch size & 4 & 8 \\
  Model size & 1024 & 2048 \\
  Max prompt & 1024 & 1024 \\
  Max sequence & 1024 & 1024 \\
  Datatype & fp16 & fp16 \\
  \bottomrule
  \end{tabular}
  \end{sc}
  \end{small}
  \end{center}
\end{table}

\subsection{Speedup for GPT-2 Layer with Different Sequence Lengths}

\begin{figure}[htbp]
\centering
\includegraphics[width=0.48\textwidth]{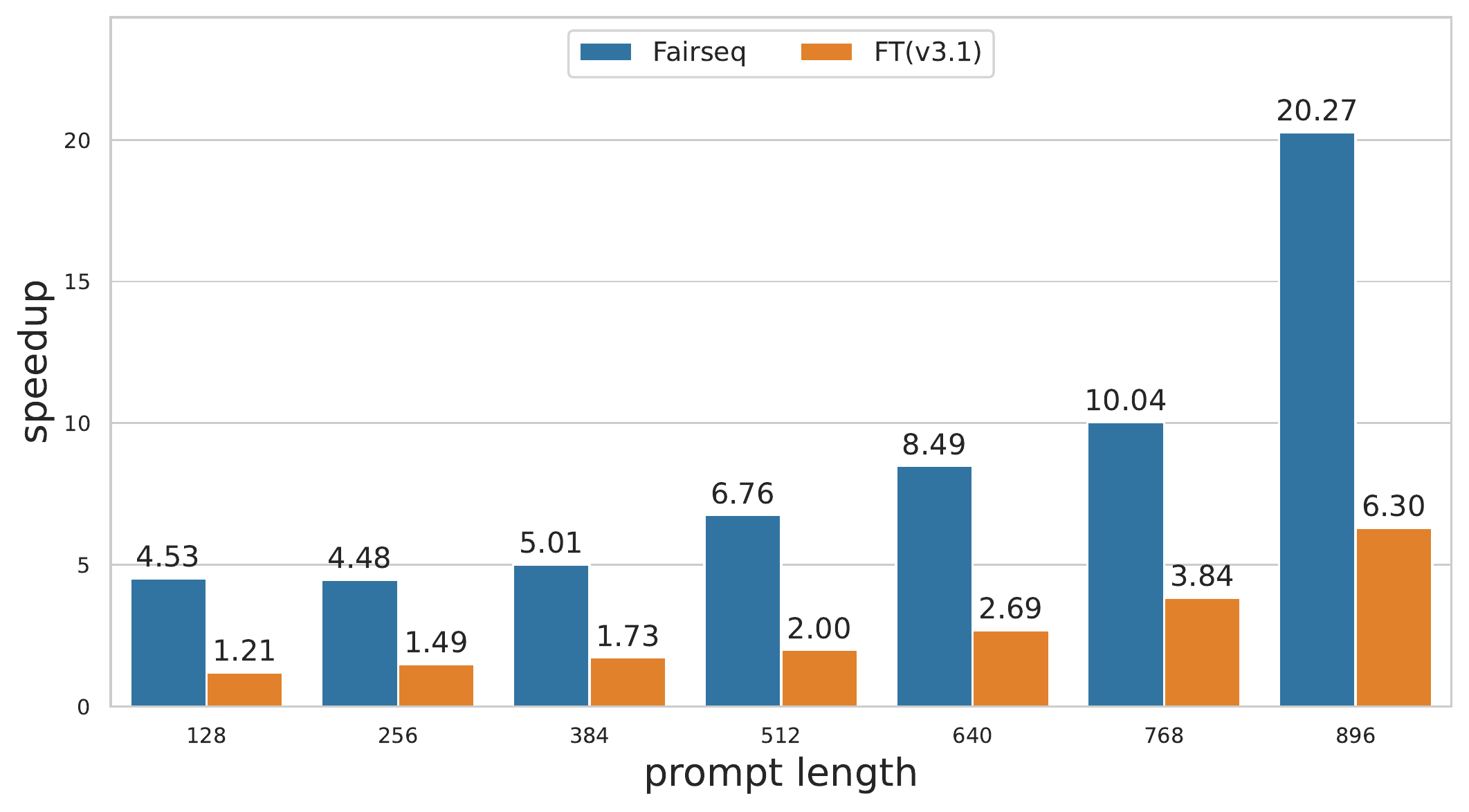}
\caption{Inference speedup of EET with different prompt lengths on 2080ti compared to Fairseq and FT(v3.1).}
\label{figure gpt2_2080}
\end{figure}

\begin{figure}[htbp]
\centering
\includegraphics[width=0.48\textwidth]{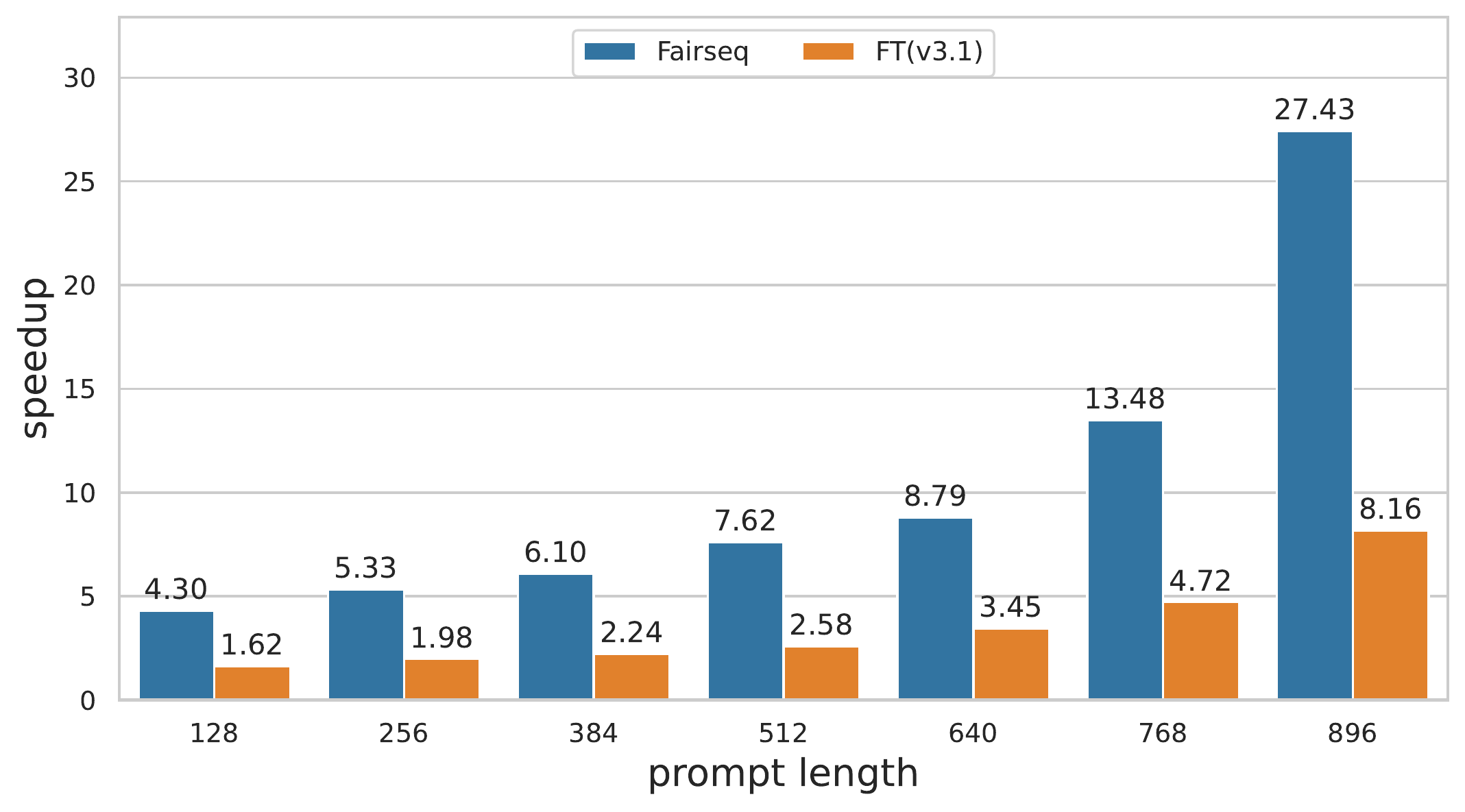}
\caption{Inference speedup of EET with different prompt lengths on A100 compared to Fairseq and FT(v3.1).}
\label{figure gpt2_a100}
\end{figure}

We first apply EET over GPT-2 on NVIDIA 2080ti and A100. Figure~\ref{figure gpt2_2080} and ~\ref{figure gpt2_a100} reveal that EET achieves about 4.48-20.27x and 4.30-27.43x speedup than Fairseq and about 1.21-6.30x and 1.62-8.16x speedup than FT(v3.1), on 2080ti and A100 respectively. For Fairseq and FT(v3.1), the incremental decoding processes the input tokens one by one, while EET improves the tokens parallelism by processing input tokens all at once. As a result, the speedup grows with the increase of the input length.

\begin{figure}[hbtp]
\centering
\includegraphics[width=0.48\textwidth]{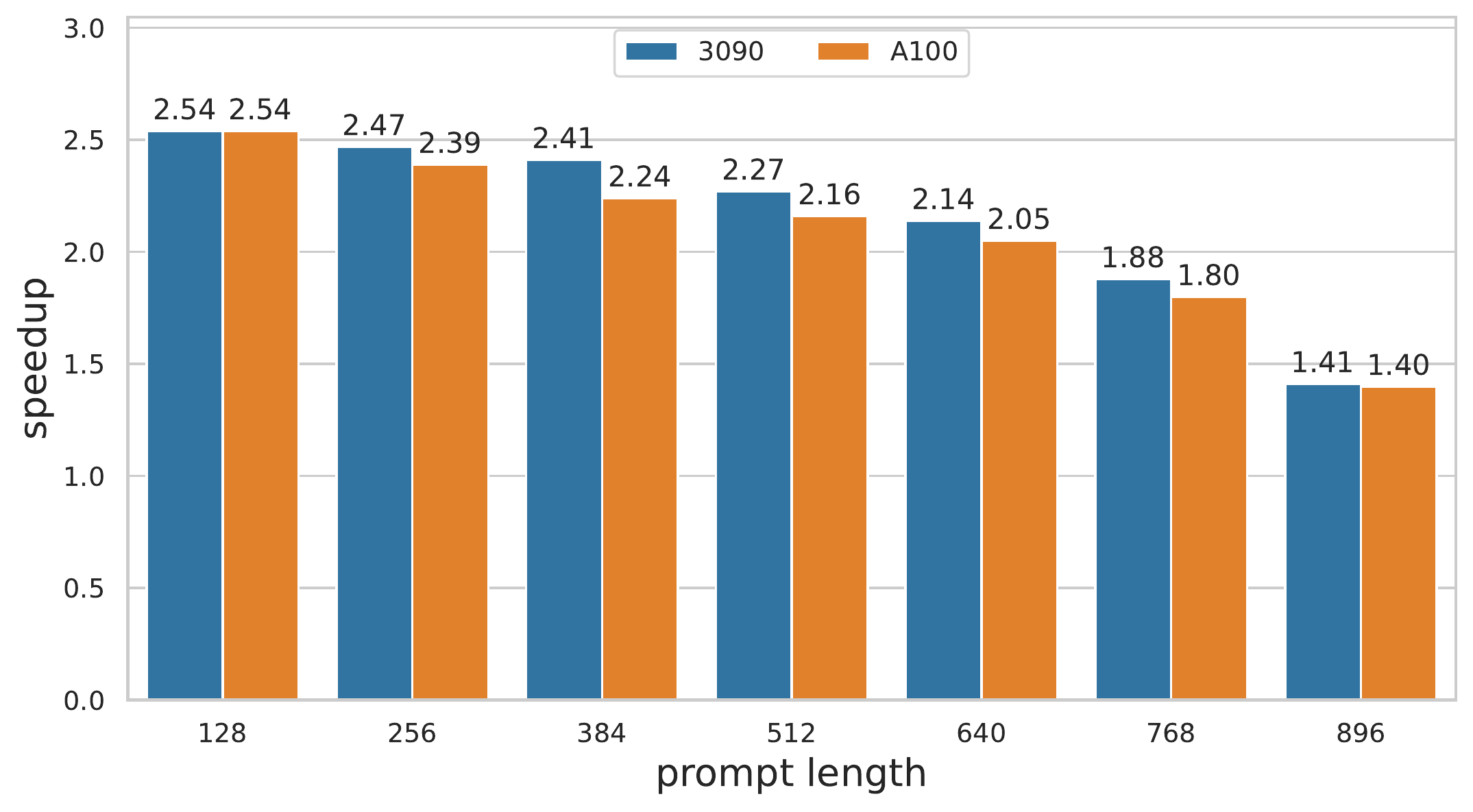}
\caption{Inference speedup of EET with different prompt lengths on A100 and 3090 over FT(v4.0).}
\label{figure gpt2_ft4.0}
\end{figure}

The recent version of FT(v4.0) also introduces the parallel decoding of the input sequences for text generation as we did, so the performance of EET and FT(v4.0) is getting closer with the input length increasing. However, EET still has some performance advantages, which are attributed to our operation kernel optimization. Figure ~\ref{figure gpt2_ft4.0} shows that EET achieves about 1.40-2.54x speedup compared to FT(v4.0) with the configuration B in Table~\ref{tabel configuration}. 

\begin{figure}[hbpt]
\centering
\includegraphics[width=0.48\textwidth]{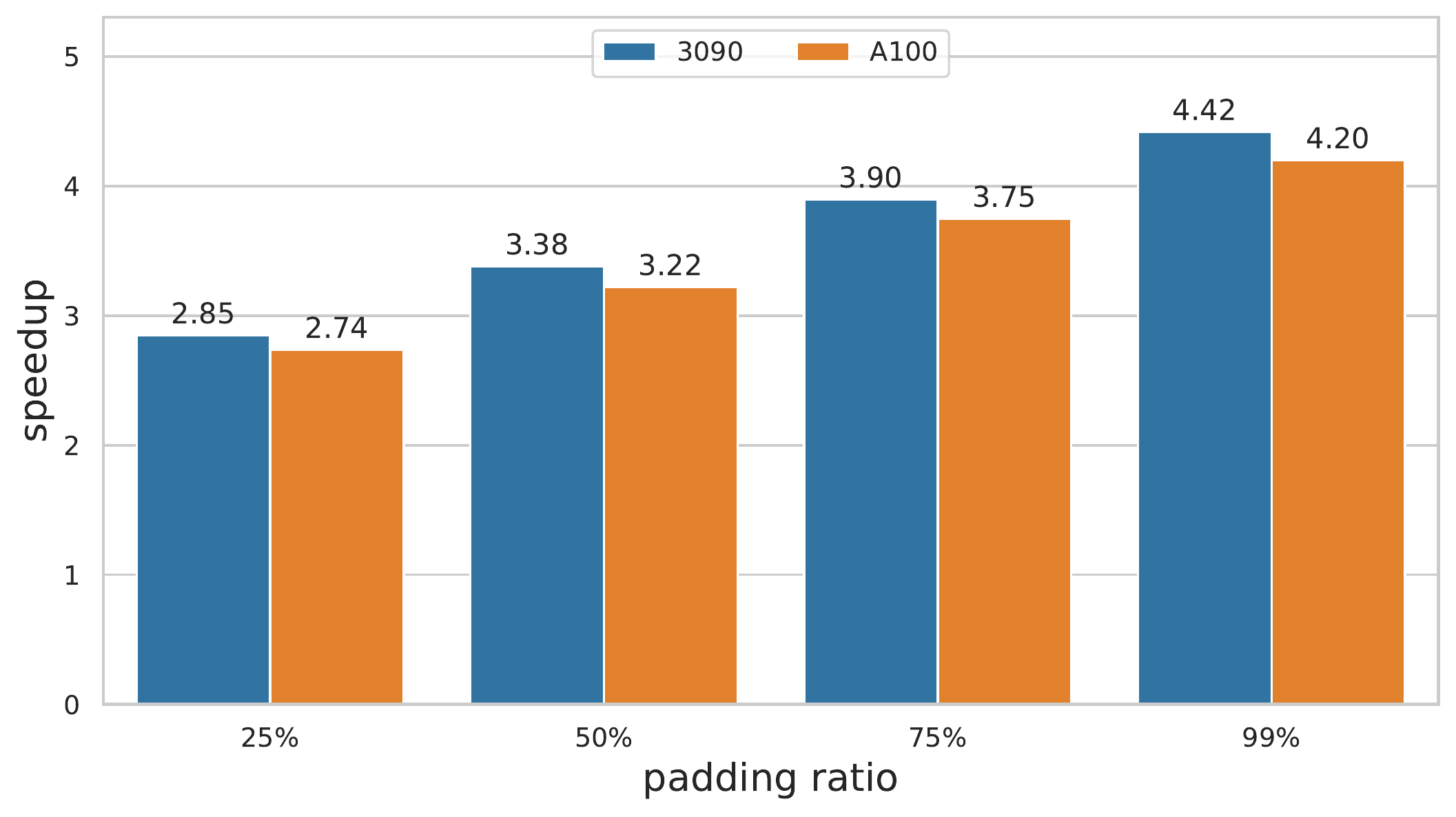}
\caption{Inference speedup of EET with different padding ratio on A100 and 3090 compared to FT(v4.0)}
\label{figure padding_ft4.0}
\end{figure}

When processing a batch of inputs, the length of them may be uneven. The FT(v4.0) uses the minimum length of the prompts for full decoding, while the EET uses the maximum length. For example, if there is a batch containing sequences of different length like [5, 2, 4, 10], the final prompt length used for parallelism is 2 in the FT. In contrast, it is 10 in the EET. Figure~\ref{figure padding_ft4.0} shows that we make 2.74-4.42x speedup with the prompt fixed to 512 and other configurations keeping the same as the configuration B in Table~\ref{tabel configuration}. 

\begin{figure}[htbp]
  \centering
  \includegraphics[width=0.48\textwidth]{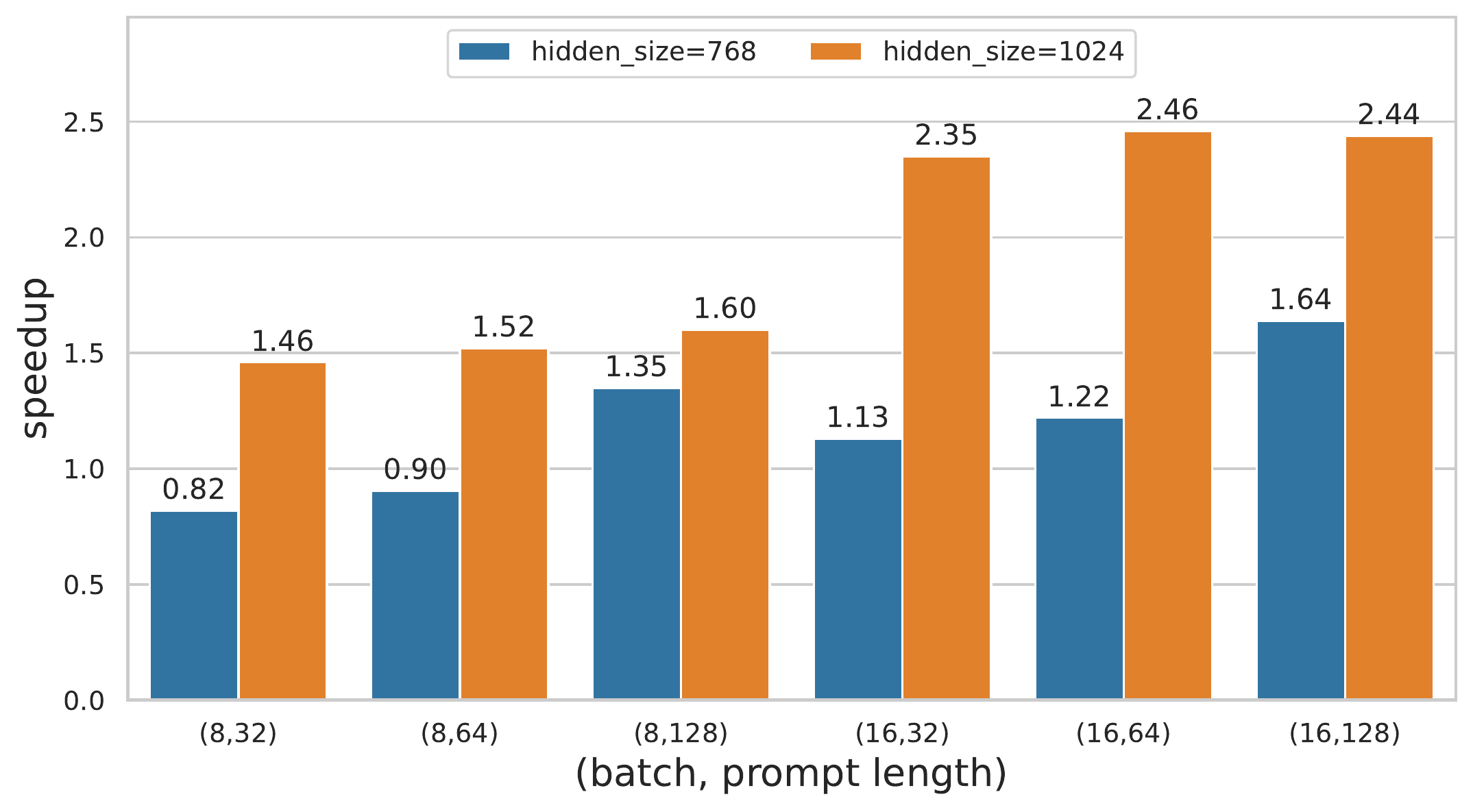}
  \caption{Speedup compared to LightSeq.}
  \label{figure LightSeq}
  \end{figure}

Unlike Fairseq and FT(v4.0), LightSeq only supports model sizes that are smaller than 1024, we also make a comparison here as a supplement. Figure~\ref{figure LightSeq} shows that we make 0.82-2.46x speedup when we set the model size to 768 and 1024. 

\subsection{Speedup for Transformer Decoder Layer with Different Model Sizes}

\begin{figure}[htbp]
\centering
\includegraphics[width=0.48\textwidth]{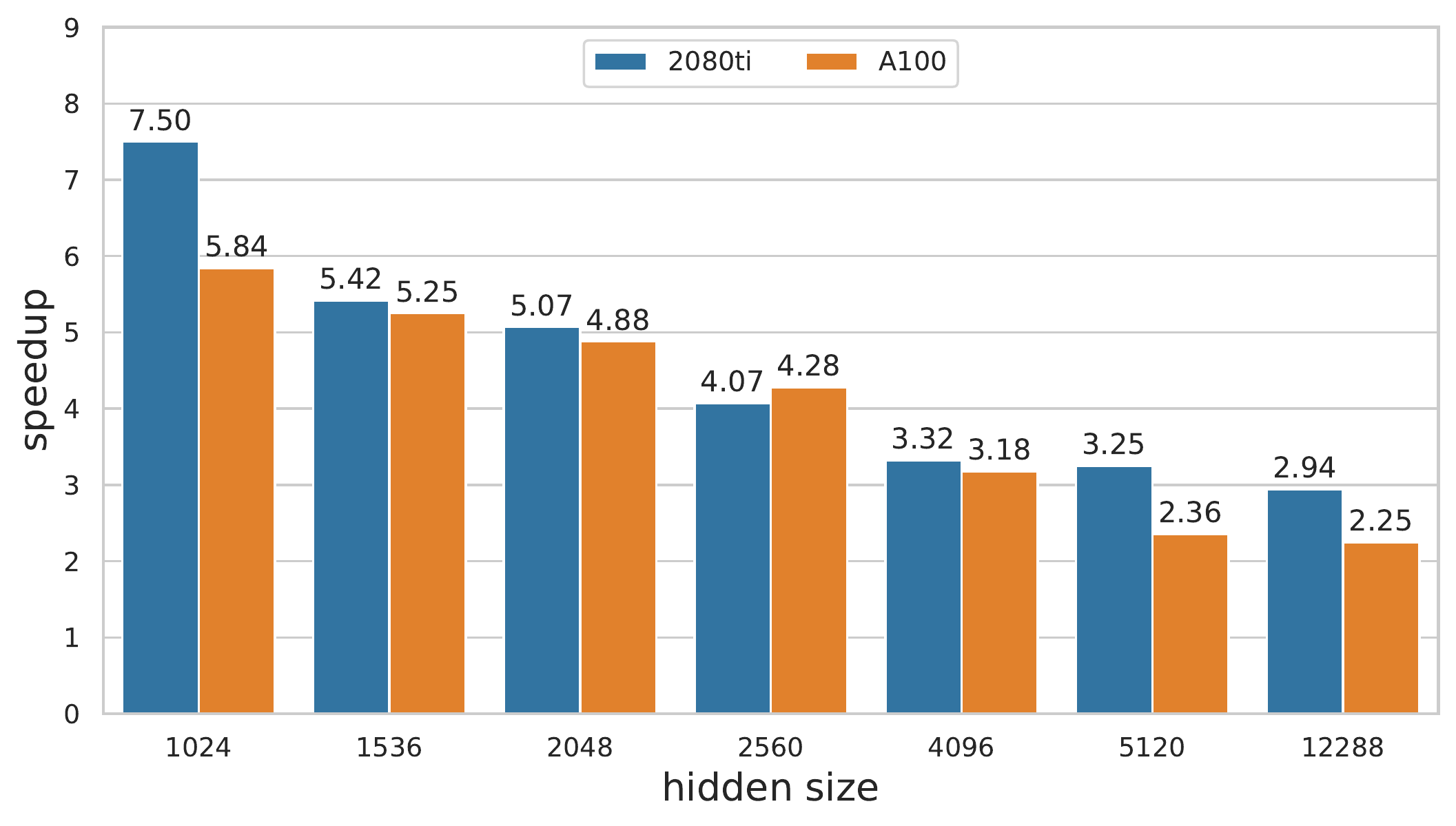}
\caption{Speedup with different model sizes on 2080ti and A100 compared to Fairseq.}
\label{figure 13}
\end{figure}

\begin{figure}[htp]
\centering
\includegraphics[width=0.48\textwidth]{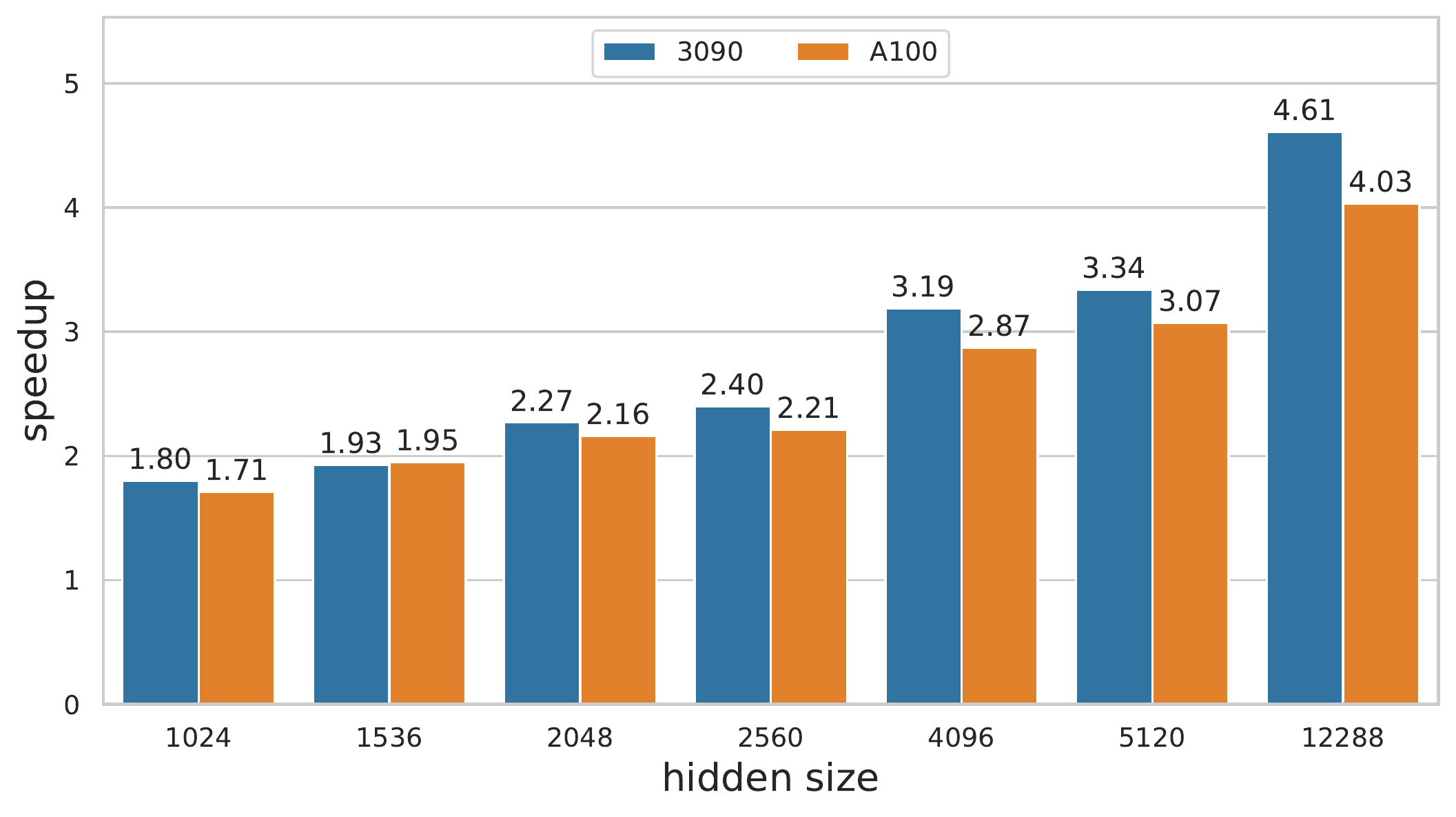}
\caption{Speedup with different hidden sizes compared to FT(v4.0).}
\label{figure 14}
\end{figure}

To prove the scalability of our EET, we evaluate the performance on different model sizes with configuration C in Table~\ref{tabel configurationC}. Figure~\ref{figure 13} and Figure~\ref{figure 14} reveal that EET achieves about 2.25-7.50x speedup than Fairseq and about 1.71-4.61x speedup than FT(v4.0). The acceleration ratio decreases as the model size increases due to the increased ratio of matrix multiplication in the inference. Nevertheless, with the help of \textit{thread block folding}, EET can still deliver significant speedup with very large model sizes, compared to Fairseq and FT(v4.0). 

\begin{table}[htbp]
  \caption{Configuration C}
  \label{tabel configurationC}
  \begin{center}
  \begin{small}
  \begin{sc}
  \begin{tabular}{lcccr}
  \toprule
   & Config C\\
  \midrule
  Batch size & 4 / 8 \\
  prompt & 512  \\
  Max sequence & 1024 \\
  Datatype & fp16 \\
  \bottomrule
  \end{tabular}
  \end{sc}
  \end{small}
  \end{center}
\end{table}
  
\subsection{Speedup for Bert Layer on 2080ti}
\begin{figure}[H]
\centering
\includegraphics[width=0.48\textwidth]{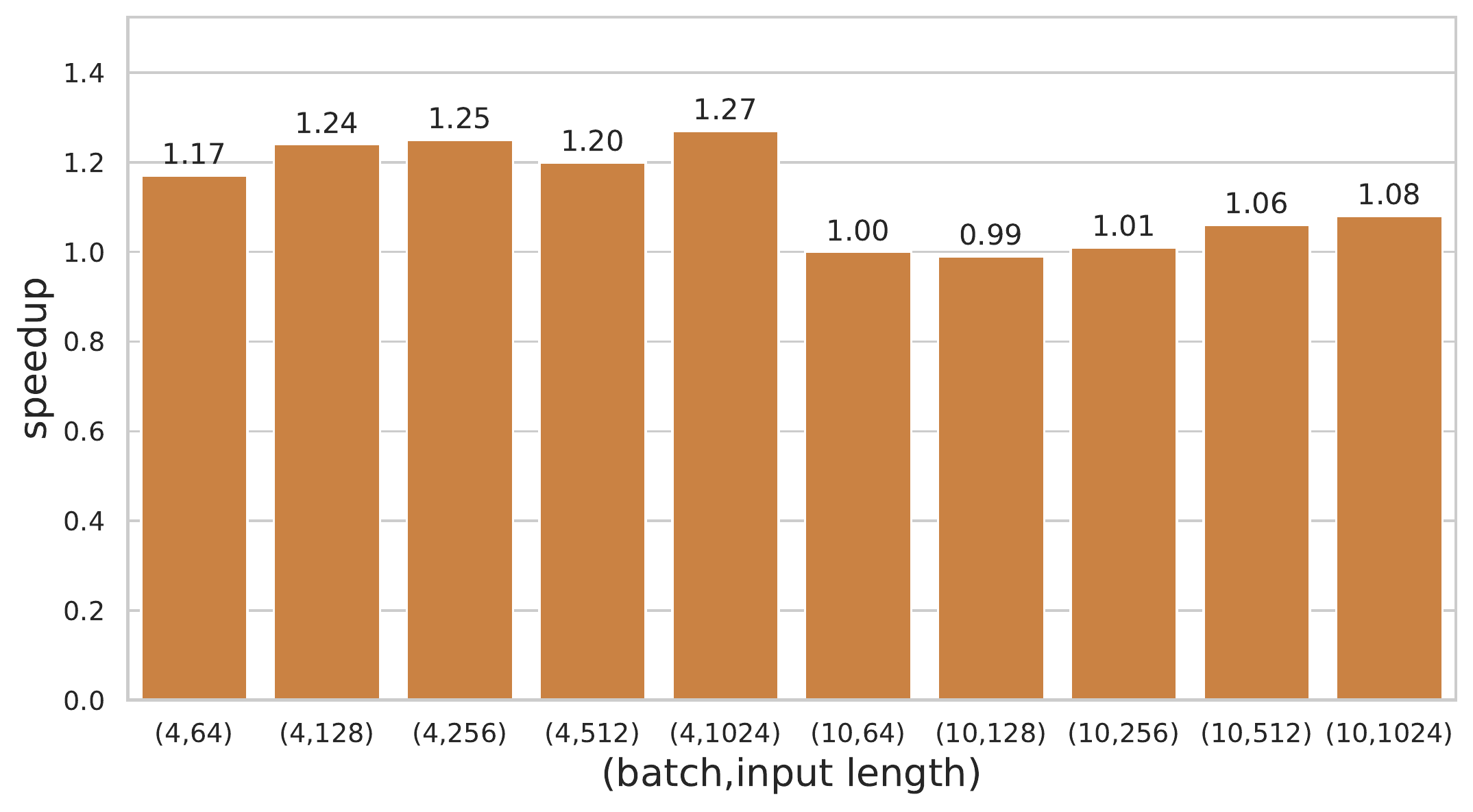}
\caption{Performance speedup for Bert layer on 2080ti compared to FT(v4.0).}
\label{figure 17}
\end{figure}

We conduct experiments to validate the performance of the Bert encoder layer in EET on 2080ti. It is worth noting that the padding tokens take up half of the total tokens. The result is shown in Figure~\ref{figure 17}. Deprecation of the padding masks with the \textit{mask fusion} trick brings in 0.99-1.27x speedup. As for Bert, its hidden size is fixed to 1024 and it has no sequence mask, which kicks off the optimization of thread-block folding and sequence mask fusion, then the speedup is not as significant as GPT2. 

\subsection{Memory distribution}
\begin{figure}[ht]
\centering
\includegraphics[width=0.48\textwidth]{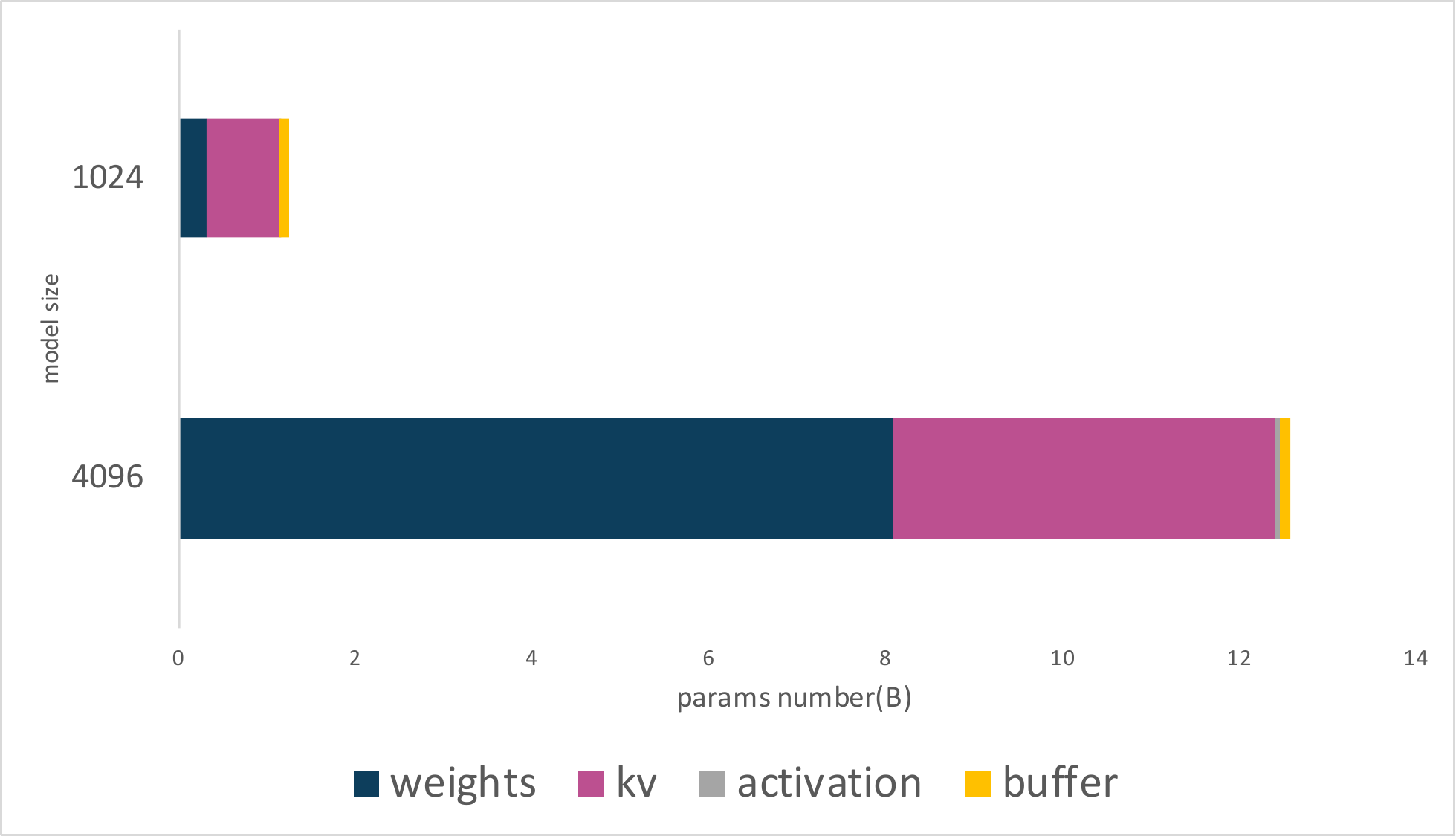}
\caption{Memory distribution for 1024/4096 hidden sizes.}
\label{figure memory_dist}
\end{figure}

Given the batch size 16, the maximum sequence length 1024, the vocab size 13672, we plot the memory distribution of the hidden size of 1024 and 4096 with layer numbers 24 and 40 respectively, as shown in Figure~\ref{figure memory_dist}. Regardless of the hidden size, we can find that model weights and $K$/$V$ caches occupy most memory. The activation caches and the buffers only take up a small part, which shows the effectiveness of our dynamic memory management strategy.

\begin{figure}[ht]
\centering
\includegraphics[width=0.48\textwidth]{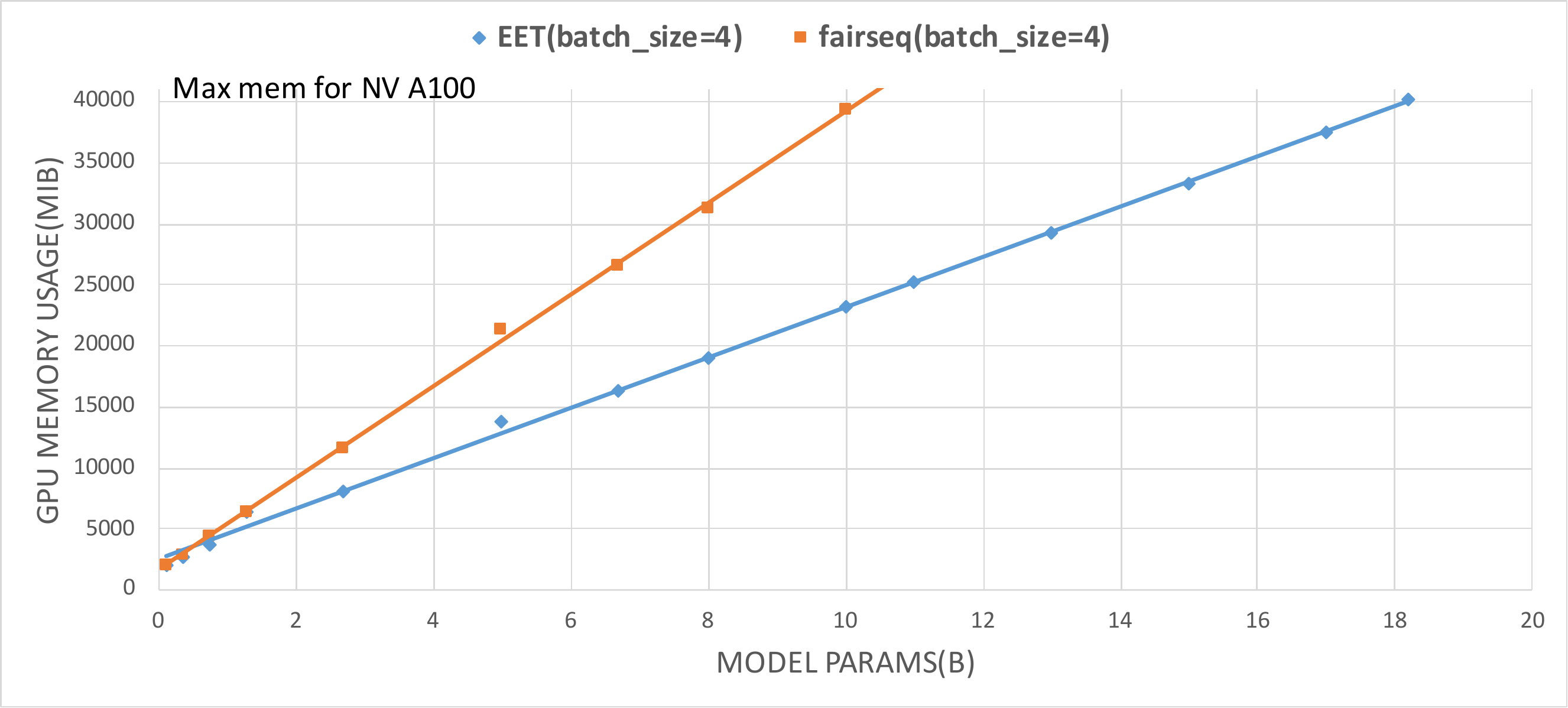}
\caption{Memory occupancy for different model sizes.}
\label{figure memory_occupancy}
\end{figure}

Given the batch size 4, the maximum sequence length 1024, we plot the memory occupancy of different model parameter sizes, see Figure~\ref{figure memory_occupancy}. Compared with the 10 billion of PyTorch's maximum model parameter sizes, it is up to 18 billion for our EET, which proves that we can place much larger models onto one GPU, thus avoiding unnecessary waste of GPU resources and inter-GPU communication overhead on multiple cards.

\section{Conclusion}
This paper comprehensively describes a series of optimization techniques for transformer inference acceleration exploiting both algorithmic and GPU hardware features. These techniques are packed into the EET, a library dedicated to inference acceleration for large transformer-based models and long input lengths. EET has a 1.40-4.42x speedup for the GPT-2 layer and a 0.99-1.27x speedup for the Bert layer compared to the state-of-the-art transformer inference library FT. To make EET easier to apply to a specific task, we provide operation level and model level API, meanwhile integrating web service with dynamic batching. We will continue to improve and keep it up-to-date.

\paragraph{Acknowledgments}
We would like to thank all the users of EET and the anonymous reviewers for their excellent feedback. This work is supported by the Key Research and Development Program of Zhejiang Province (No. 2022C01011).

\bibliography{anthology,custom}
\bibliographystyle{acl_natbib}
\end{document}